\pdfoutput=1

\documentclass[11pt]{article}

\usepackage[preprint]{acl}

\usepackage{times}
\usepackage{latexsym}
\usepackage[normalem]{ulem} 
\usepackage{color}         

\usepackage[T1]{fontenc}

\usepackage[utf8]{inputenc}

\usepackage{microtype}

\usepackage{inconsolata}

\usepackage{graphicx}
\usepackage{slnotation}
\usepackage{booktabs}
\usepackage{enumitem}
\usepackage[dvipsnames]{xcolor}
\usepackage{subcaption}
\title{Characterizing Cultural Localization in AI-Generated Stories}

\author{
    Shaily Bhatt\thanks{Equal contribution.}  \\
    Carnegie Mellon University \\
    \texttt{shaily@cmu.edu}\\
  \And
    Supriti Vijay\footnotemark[1]  \\
    Carnegie Mellon University \\
    \texttt{supritiv@cs.cmu.edu}  \\  
  \AND
    Jeremiah Milbauer  \\
    Carnegie Mellon University \\
    \texttt{jmilbaue@cs.cmu.edu}  \\  
  \And
    Fernando Diaz  \\
    Carnegie Mellon University \\
    \texttt{diazf@acm.org}  \\  
}

\usepackage{xspace}
\usepackage{subscript}

\newcommand{\word}[0]{\sWord}

\newcommand{\cultures}[0]{\slSet{C}}

\newcommand{\culture}[0]{c}

\newcommand{\topic}[0]{t}

\newcommand{\topics}[0]{\slSet{T}}

\newcommand{\story}[0]{s}
\newcommand{\storytc}[2]{\story_{#1, #2}}
\newcommand{\stories}[0]{S}

\newcommand{\lexicon}[0]{\sVocab}

\newcommand{\lexiconck}[1]{\lexicon_{#1}^{k}}

\newcommand{\stereotypesc}[1]{\slSet{Z}_{#1}}

\newcommand{\maskedstory}[0]{\overline{\story}}
\newcommand{\maskedstorytc}[2]{\maskedstory_{#1, #2}^k}
\newcommand{\maskedstories}[0]{\overline{\stories}}

\newcommand{\Fone}[0]{F\textsubscript{1}\xspace}

\newcommand{\request}[1]{[#1]\xspace}
\newcommand{\constraint}[1]{\emph{#1}\xspace}
\newcommand{\markers}[1]{\{#1\}\xspace}
\newcommand{\generation}[1]{[#1]\xspace}
\newcommand{\mask}[0]{\constraint{mask}}

\newlist{inlinelist}{enumerate*}{1}
\setlist*[inlinelist,1]{%
  label=(\alph*),
}

\definecolor{randomred}{HTML}{B91E1E}
\definecolor{unmaskedgreen}{HTML}{1A7D29}
\definecolor{cultureblue}{HTML}{2225A9}
\definecolor{baselinegray}{HTML}{2C2B2B}

\newlist{hypotheses}{enumerate}{1}
\setlist*[hypotheses,1]{label=H\arabic*:,ref=H\arabic*}

\newlist{questions}{enumerate}{1}
\setlist*[questions,1]{label=RQ\arabic*:,ref=RQ\arabic*}

\begin{document}
\maketitle

\begin{abstract}
The global use of artificial intelligence has increased interest in assessing the ability to generate culturally localized content, including stories. 
Cultural localization in stories often occurs through either templated localization---the use of cultural markers (e.g., names, locations) in a generic narrative---or holistic localization---the variation of  plots, values, and themes, in addition to cultural markers.
We propose a method to measure the degree to which content was generated through templated localization. Specifically, we identify the lexical tokens that distinguish stories across nationalities and measure the similarity of the narratives that remain after removing them.
In stories generated by five models on 125 topics for 193 nationalities, our method is able to detect that only a small subset (9-17\%) of the vocabulary accounts for the variation across nationalities
and that the narratives that remain after removing them contain repeated multi-word sequences, suggesting the presence of a shared culturally-agnostic narrative template.
Finally, we characterize the cultural markers for their stereotypicality and offensiveness, finding that markers from 19 countries, mostly located in the Global South, are on average offensive.
\end{abstract}

\section{Introduction}

Large language models (LLMs) are increasingly being used globally, requiring them to tailor their generations to diverse sociocultural contexts when instructed. For example, when given the instruction \request{Write a story about `honesty' for an Indian kid.}, a model must generate a narrative that is localized to the Indian cultural context.

\begin{figure}
    \centering
    \scriptsize
    \begin{subfigure}{0.9\columnwidth}
    
      \input{artifacts/stories/templated.tex}
      \caption{Templated localization. }
    \end{subfigure}

    \vspace{1em}

    \begin{subfigure}{0.9\columnwidth}
      \input{artifacts/stories/american-holistic}

      \vspace{1em}
      
      \input{artifacts/stories/indian-holistic}
      \caption{Holistic localization. }
    \end{subfigure}

    \caption{Example stories for the instruction \request{Write a story about `honesty' for a \constraint{nationality} kid} where \constraint{nationality} is \markers{Indian, American}.}
    \label{fig:stories}
\end{figure}

Narrative localization can take many forms, including culturally specific plot tropes \cite{colby_partial_1973}, the encoding of particular values \cite{hobson-etal-2024-story,wu_cross-cultural_2023}, variations in narrative structure \cite{song_narrative_2017}, and culturally relevant entities such as names or locations \cite{bhatt-diaz-2024-extrinsic}. We consider the two forms of localization shown in Figure \ref{fig:stories}. In \textit{templated localization}, cultural markers (e.g., names, locations) are inserted into a culturally-agnostic narrative template \cite{fan-etal-2019-strategies,ford-etal-2018-importance,wiseman-etal-2018-learning,khanuja_image_2024}. On the other hand, \textit{holistic localization}  employs culturally specific plots and settings, in addition to cultural markers.

How models localize stories has broader impacts on cultural production and preservation. While templated localization may be the appropriate choice in contexts where the goal is to preserve the content while making it more relatable to the audience \cite{khanuja_image_2024,khanuja_towards_2025}, when used in the general context, the resulting stories will often reflect homogeneous narratives and values, which can lead to cultural harms like erasure \cite{qadri2025risksculturalerasurelarge,shelby_sociotechnical_2023}, imposition of western values \cite{shelby_sociotechnical_2023,bhatt_re-contextualizing_2022,sambasivan_re-imagining_2021}, or reduced creative diversity \cite{Agarwal_2025,doshi_generative_2024}. Further, model outputs may implicitly rely on a limited set of culturally associated tokens, which prior analyses have shown can reflect stereotypical cultural associations \cite{bhagat_tales_2025,rooein_biased_2025}. Therefore, tools to detect the degree of templated localization can help anticipate and avoid potential harms.  While studies have demonstrated that stories generated in the presence of cultural cues contain lexical variation \cite{bhatt-diaz-2024-extrinsic}, misrepresentations \cite{bhagat_tales_2025}, and geographical disparity \cite{bhagat2024richeroutputrichercountries}, they have yet to provide methods for understanding the presence of templated localization.

We propose a two-stage method to detect the presence of templated localization in model generations. First, we identify the set of lexical items that function as unique cultural markers for each cultural identity in the generated stories. Next, we measure the homogeneity of the text sequences that remain after these cultural markers have been removed using multi-word similarity. Convergence of these remaining sequences across stories that differ in cultural markers would indicate the presence of a shared culturally-agnostic narrative template. We further characterize the stereotypicality and offensiveness of the cultural markers using the SeeGULL dataset \cite{jha2023seegullstereotypebenchmarkbroad}. We evaluate stories generated by five LLMs for 193 cultural identities, operationalized through nationalities, for 125 story topics curated from prior work \cite{bhatt-diaz-2024-extrinsic} and established frameworks of variation in cultural values like World Values Survey \cite{wvs2022} and Hofstede's Cultural Dimensions \cite{hofstede_values_nodate}. Our code and data is \href{https://github.com/shaily99/templated_localization}{publicly available}.

Our method reveals that localization primarily occurs through surface-level lexical differences, suggesting that stories may use a homogeneous underlying narrative. We find that cultural markers, constituting only 9-17\% of the vocabulary across models, are the only distinguishing characteristics of the stories. Moreover, the narratives that remain after removing these markers exhibit higher multi-token similarity across nationalities than the original stories. Finally, we find that countries for which cultural markers are, on average, offensive are mostly located in the Global South, predominantly in Africa and West Asia, with dominant languages that are lower-resourced. Taken together, our findings demonstrate the ability of our method to characterize cultural localization in AI-generated stories. 
\section{Background}

Narrative generation systems often decompose stories into two levels: a structural plan describing events and character relationships, and a surface realization that renders this plan into natural language. Although many early natural language generation systems used slot-filling approaches to populate manually written templates \citep{reiter:nlg,van-deemter-etal-2005-squibs}, learning-based approaches either automatically select \cite{zhou_template-filtered_2004} or generate  \cite{ford-etal-2018-importance,wiseman-etal-2018-learning,fabbri_template-based_2020,gangadharaiah_recursive_2020} templates. 
Such plan-based systems generate a story in multiple steps, including generating templates of plot and character actions, followed by rendering these plans into natural language. 
Methods of narrative planning have ranged from story grammars \cite{pemberton_modular_1989,ryan_grimes_2017} to symbolic planners \cite{riedl_narrative_2010,mcintyre_plot_2010,mcintyre_learning_2009}, and finally, to neural models \cite{martin_event_2018,xu_skeleton-based_2018,yao_plan-and-write_2019,goldfarb-tarrant_content_2020}. 
Narrative planning has also been integrated into prompt-based story generation from LLMs \cite{xie_creating_2024,li_storyler_2025}. 
Narratives have also been studied computationally by decomposing them into their attributes such as the setting, agents, events, and so on \cite{piper_narrative_2021,hamilton_narrabench_2026}.
This separation between narrative structure and its linguistic realization suggests that variation in generated stories can arise either through modifications in the underlying template or the lexical content used to instantiate it.

This distinction between structural plans and surface realization parallels theories of language variation across cultural identities. Sociolinguistic scholars have argued that social meaning and identity can be conveyed, constructed, and interpreted through various channels, including
\begin{inlinelist}
    \item micro-linguistic structures such as phonetic sounds or lexical choices,
    \item macro-linguistic forms like narrative forms or discursive orientations such as stance,
    \item entire linguistic systems such as the choice of language or dialect, and even
    \item material styles such as the choice of clothing
\end{inlinelist}
\cite{eckert_three_2012,eckert_variation_2008,bucholtz_identity_2005}. 
Importantly, these channels include narrative form such as the stance. 
Scholars of folk narrative have shown that plot and character tropes are culturally specific: the narrative structure of Russian folktales differs systematically from that of North Alaskan stories \cite{colby_partial_1973}, and similar differences have been documented in stories across other traditions \cite{polti1916dramatic,song_narrative_2017,hobson-etal-2024-story,wu_cross-cultural_2023}. A story can therefore represent cultural identity through the use of culturally-specific entities---the names, places, and objects within it---or through differences in the narrative itself. 
Since narrative generation can separate structure from surface realization, and cultural identity can be encoded at both levels, then cultural localization in generated stories could occur either through surface markers or through narrative differences.

As LLMs are deployed globally, a growing body of work has investigated their cultural competence---that is, their ability to generate outputs that reflect culturally specific knowledge, norms, and values.  While intrinsic evaluations of cultural competence focus on the ability to recall cultural values \cite{durmus2023towards,masoud_cultural_2023,alkhamissi-etal-2024-investigating,ramezani2023knowledge}, norms \cite{dwivedi2023eticor,rao_normad_2024}, artifacts \cite{seth2024dosa}, and knowledge \cite{li2024culture,singh_global_2024,maji_sanskriti_2025,sahoo_diwali_2025,chang_global_2025,myung_blend_2024}, extrinsic evaluations focus on user-facing generative tasks \cite{bhatt-diaz-2024-extrinsic,sparck_jones_evaluating_nodate}. Prior work has examined the content produced for diverse cultural identities in extrinsic tasks such as open-ended question answering, story generation, scientific writing, creating travel itineraries, and writing assistance, finding that the cultural knowledge of LLMs may not always be reflected in generative settings \cite{bhatt-diaz-2024-extrinsic,bhagat_tales_2025}; cultural representation is often stereotypical or misrepresentative \cite{rooein_biased_2025,bhagat_tales_2025,bhagat2024richeroutputrichercountries}; and generations do not adhere to expected cultural writing styles \cite{Agarwal_2025,bhatt_research_2025}. While this work demonstrates that LLMs can incorporate culturally salient tokens, it remains unclear whether the content reflects narrative differences beyond surface-level lexical variation.

Independent of cultural evaluation, recent studies have shown that LLM-generated text often exhibits substantial homogeneity across outputs.  Prior work has evaluated homogeneity of generated outputs along various dimensions including at syntactic, semantic, and narrative levels. Specifically, LLMs have been shown to generate recurring  syntactic patterns, semantically similar concepts, homogeneous discourse structures, and epistemic claims  \cite{shaib2024detectionmeasurementsyntactictemplates,sourati2025shrinkinglandscapelinguisticdiversity,wang_generalization_2025,jiang_artificial_2025,wright_epistemic_2025,namuduri_qudsim_2025}.
Finally, both qualitative and quantitative studies of LLM-generated stories find a lack of plot diversity, recurring narrative themes, lack in pacing and tension, and positive endings  \cite{xu2024echoesaiquantifyinglack,tian-etal-2024-large-language,Begus2024,priyanshu2024silentcurriculumdoesllm}. If LLM outputs tend to reuse shared narrative structures, then cultural adaptation in generated stories may occur primarily through surface-level markers rather than through holistic localization.

Together, these observations suggest that cultural variation in generated stories may arise primarily through surface-level lexical markers rather than through deeper narrative differences. However, existing work has not directly examined whether stories generated across cultures exhibit templated localization, where cultural markers are inserted into culturally-agnostic narrative templates.
\section{Method}

We are interested in measuring the degree to which generated stories across cultures reflect templated or holistic localization.
We distinguish between these two as follows:

\paragraph{Templated localization} refers to localization when culture is represented through isolated lexical items. 
Here, cultural markers such as cultural artifacts, relevant names and locations, or other entities are inserted into culturally-agnostic templates that reflect homogeneous narrative structures, plots, settings, themes, and values.

\paragraph{Holistic localization} refers to localization when culture shapes the narrative. Here, cultural markers are distributed throughout the story, resulting in culturally-specific narrative structures, plots, themes, and values.

\subsection{Overview}

Given a prompt \request{Write a children's story about \constraint{topic} for a/an \constraint{nationality} kid in English.}, we are interested in understanding if a generated story is composed of:
\begin{inlinelist}
\item a culturally-agnostic template about \constraint{topic} shared across nationalities and
\item a set of cultural markers inserted into that template.
\end{inlinelist}
To do so, we fix \constraint{topic} and vary \constraint{nationality} to produce stories. 
Our method analyzes these stories in two stages. First, we identify the set of cultural markers that distinguish the stories (\S\ref{sec:methods:detecting-templated-localization}). Second, we measure the similarity of the narrative that remains after these cultural markers are removed (\S\ref{sec:methods:homogeneity}). Finally, we characterize the stereotypicality of the cultural markers (\S\ref{sec:methods-stereotype}).

To make cultural localization tractable for computational analysis, we impose several methodological constraints on the scope of our study.  First, because templated localization assumes exact repeated language across cultural contexts, we adopt lexical units (words) as our unit of analysis, allowing us to leverage existing natural language processing tools.  Second, while imperfect, nationality serves as a proxy for culture consistent with existing research \cite{adilazuarda-etal-2024-towards}, making the analysis amenable to classification methods. Finally, we restrict our analysis to English to facilitate lexical comparison, leaving cross-lingual template detection for an area of future study.

\subsection{Identifying Cultural Markers}
\label{sec:methods:detecting-templated-localization}

The first step of our method identifies the minimal set of words per nationality whose removal renders stories across cultures indistinguishable.
Under templated localization, lexical differences will be concentrated in a small number of cultural markers, whose removal will eliminate variation. 
By contrast, under holistic localization, the differences would be distributed throughout the story, requiring many words to be removed before stories converge.

\paragraph{Scoring candidate cultural markers.}  
Let $\storytc{\topic}{\culture}\in\stories$ be the generated story for topic $\topic\in\topics$ and culture $\culture\in\cultures$.   The vocabulary $\lexicon$ is the union of words present across all stories.  For each $\culture\in\cultures$, we score every word $\word\in\lexicon$ according to its normalized pointwise mutual information (NPMI) with $\culture$ (Appendix \ref{app:cultural-markers}).  We refer to these scored words as the candidate cultural markers of $\culture$.

\paragraph{Identifying distinguishing cultural markers.} 
Given the candidate cultural markers for $\culture$, the final set of cultural markers for $\culture$ is $\lexiconck{\culture}\subset\lexicon$,  composed of the top $k\%$ candidates with highest NPMI values.  
Let $\maskedstorytc{\topic}{\culture}$ be the story $\storytc{\topic}{\culture}$ with $\lexiconck{\culture}$ removed. 
In order to determine $k$, we measure the ability of a classifier to identify the culture of $\maskedstorytc{\topic}{\culture}$ amongst the set $\maskedstories_\topic^k=\{\maskedstorytc{\topic}{\culture}\}_{\culture\in\cultures}$.  
Specifically, at varying values of $k$, we record the \Fone of the classifier. If a subset of the vocabulary is the only identifiable characteristic of the stories across cultures, then masking words with high cultural association should make the stories indistinguishable. While the performance of the classifier should drop more significantly when words with higher cultural association are masked, our measurement question is how many culturally-associated words need to be removed. 
We refer to $\lexiconck{\culture}$ as the subset of words whose removal makes the stories in $\maskedstories_\topic^k$ indistinguishable.  We refer to the resulting stories in $\maskedstories_\topic^k$ as the template images.

\subsection{Homogeneity of Remaining Narratives}
\label{sec:methods:homogeneity}

The second step of our method detects the presence of a shared generic narrative template by measuring the homogeneity of the template images remaining after removing the cultural markers.

Although template images are indistinguishable by construction, we need a method to determine whether this is due to randomness or homogeneity amongst images.
We can measure the homogeneity of template images by computing the pair-wise average similarity amongst elements. Such measures have been used in prior work on  measuring homogeneity in a corpus \cite{padmakumar2024doeswritinglanguagemodels,shaib-etal-2025-standardizing}.  
If stories reuse a template, replacing cultural markers with masked tokens would make the resulting text sequences more similar, as compared to the original stories. 
Consider the two stories from the two cultures as \generation{A cat sat on the table.} and \generation{A dog sat on the floor.}.
Let \markers{cat, table} and \markers{dog, floor} be the markers of the respective cultures. Then, masking these markers will produce the same n-gram sequence \generation{A \mask{} sat on the \mask{}}, resulting in higher similarity compared to the original stories, as well as stories where random words are masked. 

While this stylized demonstration of homogeneity suggests that template images are exact duplicates, in practice, due to model stochasticity, we need similarity metrics robust to small perturbations amongst template images. To do so, we adopt two metrics for analyzing multi-word sequences.
In the first, we calculate the length of common substring (LCS), normalized by the length of the stories. In the second, we measure the similarity between the sets of $n$-grams present in pairs of template images, using Jaccard similarity. 
This method has been used to robustly detect duplicates in large corpora such as web crawls \cite{broder_syntactic_1997}.

Since our goal is to measure whether template images are shared across cultures, multi-word similarity offers a relatively simple yet efficient method to compare pairs of text sequences that remain after removing cultural markers, unlike other representations like discourse structures, narrative components, or themes that require more manual or computational effort \cite{namuduri_qudsim_2025,Begus2024,piper_narrative_2021}.

\subsection{Characterizing Cultural Markers}
\label{sec:methods-stereotype}
Finally, we characterize cultural markers for their degree of stereotypicality. Assume  we have access to a set of stereotypical attributes for $\culture$,  denoted as $\stereotypesc{\culture}$.  We  calculate the overlap between $\stereotypesc{\culture}$ and $\lexiconck{\culture}$ by  measuring the precision of stereotypes in the cultural markers (Appendix \ref{app:stereotype-method}).

\section{Experimental Materials}

\paragraph{Story Topics.}
We curate a set of 125 story topics that consists of 35 topics from prior work \cite{bhatt-diaz-2024-extrinsic}, and 90 based on the World Value Survey \cite{wvs2022}, Hofstede's cultural dimensions \cite{hofstede_values_nodate}, and the Moral Foundations Theory \cite{Graham2012MoralFT}. We select these frameworks as they are known to capture variation in values across cultures 
and have been utilized to evaluate AI systems' knowledge of cultural values \cite{durmus2023towards,masoud_cultural_2023}.
To curate this list, two authors read and discussed each dimension of three theories and distilled them into a topic. For example, question Q110 from the World Value Survey about rating the amount of corruption in the country is distilled into the topic `corruption'. Similarly, question Q03 from Hofstede's survey on the importance of getting recognition for good performance in the workplace is distilled into `recognition.' The complete list of topics and their corresponding sources is available in our \href{https://github.com/shaily99/templated_localization}{data}.

\paragraph{Prompts.}
We use a simple prompt template, \request{Write a children's story about \constraint{topic} for a/an \constraint{nationality} kid in English.}.
Similar to prior works, we opt for a simple instruction to generate a story  \cite{rooein_biased_2025,bhagat_tales_2025,bhatt-diaz-2024-extrinsic}, leaving examination of localization behavior in other user interaction patterns and domains to future work. 
We generate prompts for each of the 125 topics for all of the 193 nationalities, resulting in 24,125 prompts.

\paragraph{Models.}
We generate stories from two closed-source models---GPT 3.5 Turbo and GPT 4o Mini queried through OpenAI API in June 2025---and three open-weights models of varying sizes---Llama 3.1 8B Instruct, Llama 3.3 70B Instruct \cite{grattafiori_llama_2024}, and Gemma 3 12B Instruct \cite{team_gemma_2025} hosted locally using vLLM with 8-bit quantization. 
This selection balances recency, size, and open-source availability, demonstrating the effectiveness of our method across a range of models.
For all models, we set the temperature to 0.7 and the maximum tokens to 1000. To account for non-determinism during generation, we sample five responses per prompt. This results in 120,625 stories from each model.

\begin{table*}[]
\centering
\footnotesize
\begin{tabular}{lll}
\toprule
Template image                                                         & \# cultures & Example cultural markers                                                                                \\ \midrule
there lived a young boy named \mask & 193 & (America: timmy),  (China: liwei), (France: julien), (India: arjun)                                   \\
                                         &     &                                                                                        \\
\mask became a role model           & 192 & (America: charlie), (India: aarav), (Japan: haruto), (Brazil: lucas)                   \\
                                         &     &                                                                                        \\

school called \mask academy /       & 163 & (Canada: maplewood), (Bangladesh: shikha), (Japan: sakura),                            \\
school called \mask elementary      &     & (Slovakia: hrdinova), (South Korea: hanbok)                                            \\
                                         &     &                                                                                        \\
loved to play \mask /               & 58  & (America: baseball), (India: cricket), (Canada: hockey),                               \\ 
loved playing \mask                 &     & (Brazil: soccer), (Jamaica: football)                                                  \\\bottomrule
\end{tabular}
\caption{Example of Template images, number of cultures they were found in, and respective cultural markers.}
\label{tab:template-examples}
\end{table*}

\paragraph{Nationality Classifier.}
We train the nationality classifier used in Section \ref{sec:methods:detecting-templated-localization} as a multi-class (193-way)  classifier to classify stories into one of 193 nationalities. 
We fine-tune the mmBERT model \cite{marone2025mmbertmodernmultilingualencoder} using a classification head.
We use 5-fold cross-validation with a 60:20:20 split across folds for training, validation, and testing, respectively.
All the classifiers are trained for a maximum of fifty epochs, with early stopping patience set to five epochs.
The validation split is used to pick the best classifier from a combination of hyperparameters, including learning rates, batch size, and for early stopping (best parameters reported in Appendix \ref{app:classifier-details}). 
We then record the performance of the classifier on the test split. 
Additionally, we record the performance on the masked stories created from this test split.
We run the experiment independently for each of the five LLMs.

\paragraph{Template Image Similarity.}
We compare the average similarity amongst template images  with the average pair-wise similarity amongst
\begin{inlinelist}
    \item original stories, and
    \item stories when an equivalent number of random words are masked.
\end{inlinelist} 
When computing $n$-gram similarity, we use $n=4$.

\paragraph{Stereotype Data.}
In order to characterize the stereotypicality of cultural markers, we calculate the precision of stereotypes using the stereotypical attributes released in the SeeGULL dataset \cite{jha2023seegullstereotypebenchmarkbroad}. 
To create this dataset, candidate stereotypes were first sourced from language models, followed by obtaining annotations to rate the candidates as stereotypical (or not) from raters residing in the respective countries (in-group regional raters) and North American annotators (out-group raters).
We use all attributes that were labeled as a stereotype by at least one regional rater as our reference set of stereotypes ($Z_\culture$).
We present evaluation results for the 156 countries from the SeeGULL dataset that overlapped with 193 nationalities in our list. 
Further, SeeGULL provides an offensiveness score for every stereotype. Specifically, a stereotype is rated as non-offensive (-1), neutral (0), and offensive (Likert scale of 1-5), averaged across three raters. We use this to calculate the average offensiveness of stereotypical cultural markers.

\section{Results}
\subsection{Identifying Cultural Markers}
\label{sec:classifier-results}

\begin{figure}
    \centering
    \includegraphics[width=\linewidth]{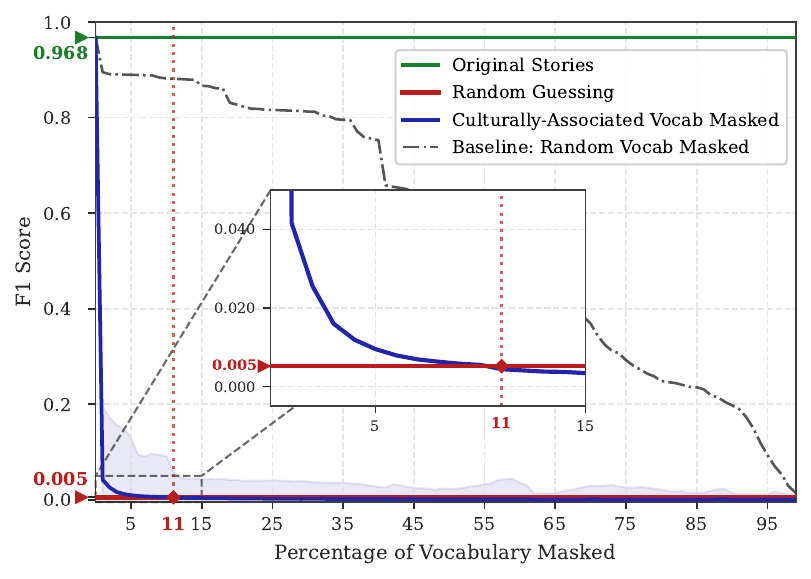}
    \caption{Identifying cultural markers. \Fone of nationality classifier as a function of number of masked words.  Results for GPT 4o Mini generation.  Results for other models can be found in Appendix \ref{app:classifier-results}.
    }
    \label{fig:f1_gpt4o}
\end{figure}

Figure \ref{fig:f1_gpt4o} shows the \Fone of the nationality classifier for all integer values of $k$ between 0 and 99 for stories generated by GPT 4o Mini. The \Fone on the original, unmasked stories is 0.968, indicating that the classifier is able to reliably predict the nationality.  For reference, randomly guessing the nationality would achieve an \Fone of 0.005.

We observe that masking increasing numbers of highly culturally associated words dramatically degrades both the macro-averaged \Fone and the class-wise \Fone of the classifier, suggesting that the ability to distinguish stories is concentrated on a small number of cultural markers. More concretely, we find that the classifier performance drops to random guessing when the top 11\% of the highly associated cultural words are masked.   Results for other models (Appendix \ref{app:classifier-results}) indicate similar fractions of cultural markers: GPT 3.5 Turbo (11\%), Llama 3.3 70B Instruct (9\%), Gemma 3 12B Instruct (9\%), and Llama 3.1 8B Instruct (17\%).

To ensure that our results were not an artifact of merely removing words, we compared the \Fone to masking random words.  While the \Fone of the classifier in this condition also reduces as $k$ increases, it drops more slowly than when words ordered by cultural association are masked.  We find that, for all values of $k$, the \Fone when random words are masked is higher than that when words with the highest cultural association are masked (one-sided paired $t$-test, $p < 0.05$).

\subsection{Homogeneity of Remaining Narratives}
\label{sec:similarity-results}

We now turn to evaluating the homogeneity in template images that remain after masking cultural markers from the stories using multi-word similarity (\S~\ref{sec:methods:homogeneity}). Table \ref{tab:template-examples} shows examples of template images that were repeated across nationalities.
Table \ref{tab:homogeneity} shows the average multi-word similarity amongst the original stories and their template images.  We break similarity down into inter-group similarity---amongst stories across cultures---and intra-group similarity---amongst stories within a culture. 

For inter-group similarity, across both LCS and Jaccard, we find that similarity amongst template images is higher than amongst original stories. Appendix table \ref{tab:app-homogeneity} shows that masking an equivalent number of random words results in lower similarity than masking cultural markers. Together, these findings demonstrate that the sequences remaining after masking cultural markers contain a latent culturally-agnostic narrative template.

\begin{table*}[t]
    \centering
    \footnotesize
    \begin{tabular}{lccccccccc}
    \toprule
     & \multicolumn{3}{c}{inter-group} & \multicolumn{3}{c}{intra-group} & \multicolumn{3}{c}{inter/intra} \\
    \cmidrule(lr){2-4} \cmidrule(lr){5-7} \cmidrule(lr){8-10}
     & original & masked & \% inc. & original & masked & \% inc. & original & masked & \% inc. \\
    \midrule
    \multicolumn{10}{c}{Longest Common Subsequence} \\
    \midrule
    GPT 3.5 Turbo & 0.0252 & 0.0307 & 21.8\% & 0.0347 & 0.0379 & 9.2\% & 0.7262 & 0.8100 & 11.5\% \\
    GPT 4o Mini & 0.0135 & 0.0161 & 19.3\% & 0.0172 & 0.0188 & 9.3\% & 0.7849 & 0.8564 & 9.1\% \\
    Gemma 3 12B Instruct & 0.0114 & 0.0129 & 13.2\% & 0.0137 & 0.0152 & 10.9\% & 0.8321 & 0.8487 & 2.0\% \\
    Llama 3.1 8B Instruct & 0.0122 & 0.0136 & 11.5\% & 0.0131 & 0.0143 & 9.2\% & 0.9313 & 0.9510 & 2.1\% \\
    Llama 3.3 70B Instruct & 0.0180 & 0.0222 & 23.3\% & 0.0263 & 0.0290 & 10.3\% & 0.6844 & 0.7655 & 11.9\% \\
    \midrule
    \multicolumn{10}{c}{Jaccard (4-gram shingles)} \\
    \midrule
    GPT 3.5 Turbo & 0.0299 & 0.0376 & 25.8\% & 0.0419 & 0.0462 & 10.3\% & 0.7136 & 0.8139 & 14.0\% \\
    GPT 4o Mini & 0.0161 & 0.0206 & 28.0\% & 0.0217 & 0.0247 & 13.8\% & 0.7419 & 0.8340 & 12.4\% \\
    Gemma 3 12B Instruct & 0.0075 & 0.0125 & 66.7\% & 0.0122 & 0.0177 & 45.1\% & 0.6148 & 0.7062 & 14.9\% \\
    Llama 3.1 8B Instruct & 0.0093 & 0.0125 & 34.4\% & 0.0110 & 0.0141 & 28.2\% & 0.8455 & 0.8865 & 4.9\% \\
    Llama 3.3 70B Instruct & 0.0218 & 0.0282 & 29.4\% & 0.0339 & 0.0379 & 11.8\% & 0.6431 & 0.7441 & 15.7\% \\
    \bottomrule
    \end{tabular}
    \caption{Narrative homogeneity.  Multi-word similarity amongst stories on the same topic in either their original form or masked (Section \ref{sec:methods:homogeneity}).  
    Inter-group measures the similarity amongst stories for different nationalities.  Intra-group measures the similarity amongst stories for the same nationality.  The last column group divides the inter-group similarity by the intra-group similarity to control for similarity attributable to a cross-nationality template.  }
    \label{tab:homogeneity}
    \end{table*} 

Comparing the inter-group with the intra-group similarity, we observe a consistently higher average similarity in the latter, suggesting that, even after removing cultural markers, some signals of nationality remain in the masked stories.  While higher than inter-group similarity, the differences are modest, and we speculate that a relatively large fraction of this similarity is attributable to a generic latent template.  In the final column group, we show that inter-group similarity after masking accounts for more than 70\% of the intra-group similarity, across both measures.

\subsection{Presence of Stereotypes}
\label{sec:stereotype-results}

The top 10 countries with the highest stereotype precision and offensiveness of the cultural markers for GPT 4o Mini stories are in Table \ref{tab:stereotype-results}. 
Results for other models and examples are in Appendix \ref{app:stereotype-results}. 

The top countries with the highest stereotype precision tend to be countries where higher-resourced languages are dominant.   We find that the stereotypes present in cultural markers have varying degrees of offensiveness across countries.
In 42 of the 61 countries with non-zero stereotype precision, the average offensiveness score is negative or neutral, indicating that the stereotypical cultural markers were rated as non-offensive by the annotators.
The 19 countries with positive average offensiveness scores are primarily located in the Global South, predominantly in Africa and West Asia, with dominant languages being lower-resourced.

\begin{table}[]
\footnotesize
\begin{tabular}{lrlr}
\toprule
Nationality  & Precision & Nationality  & Offensiveness \\ \midrule
Australia    & 0.0193                        & Tunisia      & 3.667                           \\
United States      & 0.0128                        & Syria        & 3.667                           \\
India        & 0.0109                        & Serbia       & 3.000                           \\
Ethiopia     & 0.0062                        & El Salvador & 2.667                           \\
China        & 0.0060                        & Ecuador      & 2.667                           \\
New Zealand & 0.0054                        & Poland       & 2.667                           \\
Japan        & 0.0048                        & Bangladesh   & 2.667                           \\
Italy        & 0.0046                        & Georgia      & 2.667                           \\
Vietnam      & 0.0037                        & Guinea       & 2.333                           \\
North Korea  & 0.0037                        & Gambia       & 2.333                           \\ \bottomrule
\end{tabular}
\caption{Stereotypes. Top 10 countries with highest stereotype precision and offensiveness for cultural markers in stories generated by GPT 4o Mini.}
\label{tab:stereotype-results}
\end{table}

\section{Discussion}

Our work provides an analytical lens for extrinsic cultural competence by examining how models, in response to story-generation prompts, adapt narratives across cultural identities.
Our results suggest that cultural localization in AI-generated stories occurs primarily through lexical insertion of cultural markers in a culturally-agnostic template rather than through holistic changes to the narrative.  This reduces cultural representation to a small number of recognizable cultural symbols, resulting in localization underpinned by a homogeneous narrative worldview. Evaluations that do not analyze the mechanisms of cultural representation in text may overestimate AI's cultural competence.

\subsection{Localization}

The observed increase in multi-word similarity after removing cultural markers indicates the presence of latent narrative templates reused across nationalities. If differences in stories were distributed throughout the narrative (holistic localization), masking a subset of the vocabulary— or cultural markers—would not significantly increase multi-word sequence similarity as the remaining narratives would still diverge structurally. While we observe slightly higher similarity within stories from a nationality, our experiments suggest that this may largely be attributed to generic, cross-nationality templates. As a result, current LLM story generation seems to behave similarly to template-based generation pipelines despite being trained end-to-end.

Templated localization likely arises from systemic behavior resulting from LLM training.
Despite being instructed to generate narratives for varying cultural identities, models risk reverting to globally dominant narrative schemas.
This indicates that the assessment and improvement of cultural competence of AI in narrative localization needs to be broadened from the incorporation of culturally salient entities to other channels, such as narrative structures, values, stance, dialect, and so on, as suggested by the sociolinguistics literature.  

Even within the narrative channel, human raters will need specialized knowledge to make reliable assessments.
While prior work in evaluating model generations has advocated for the recruitment of participants with lived cultural experience \cite{bhagat_tales_2025,Agarwal_2025,qadri_case_2025}, non-experts and experts can differ in their judgments, for example, when evaluating the quality of machine translated text \cite{freitag_experts_2021} or success in emulating writing styles \cite{chakrabarty_can_2026}. 
Since narratives within a cultural context can vary subtly, recruitment should be done with care, potentially requiring deeper expertise with the domain (e.g., scholars). 
This echoes recent calls to develop AI in collaboration with humanities experts \citep{dagstuhl:ai-culture,hemment:doing-ai-differently,born:cifar}.  

While template-based generation does not inherently exhibit templated localization (e.g., a culture- conditioned template), we observe that cultural homogenization can surface through generic template-like behavior from LLMs which presumably respond without using explicit templates. This highlights the importance of understanding how implicit structuring (e.g., templates, plans) or explicit tool use (e.g., retrieval-augmented generation) can result in narrative homogenization. This requires developing methods for identifying and measuring homogenization throughout the reasoning and tool-use process.

\subsection{Stereotyping}
While the cultural markers for most countries contained neutral or non-offensive stereotypical attributes, the presence of offensive stereotypes for particular regions (\S\ref{sec:stereotype-results}) indicates potential for uneven representational harms.  
Further, when AI systems are used to access cultural representations of communities through tasks like narrative generation---either by members within or those outside the group---stereotypes that are neutral or non-offensive can propagate homogeneous stereotypical markers and narratives \textit{within} the community \cite{wang:stereotypes,seth_how_2025}.
This suggests the need to enrich notions of stereotypes to include \textit{narrative stereotypes}, or stereotypes in narrative structures, styles, and plots, as well as those found in other modes of cultural representation.

\section{Conclusion}
In this work, we propose a method to examine how large language models localize narratives when generating stories for different cultural contexts. 
Specifically, we assess whether localization is templated, where cultural markers are substituted into a culturally-agnostic template, as opposed to holistic localization, where cultural context shapes the narrative through differences in plot, themes, or values throughout the story. Our method first identifies the cultural markers that distinguish stories across cultural contexts and then measures the similarity of the narratives that remain after removing these markers.
Across five models, 125 topics, and 193 nationalities, we find that cultural variation is limited to a small subset of vocabulary; masking only 9-17\% of culturally-associated words renders stories across cultures indistinguishable. Moreover, many of these markers are stereotypical, and markers from 19 countries, primarily located in the Global South, are, on average, offensive, while those from the rest are non-offensive or neutral.
Further, after masking these cultural markers, the remaining narratives become more similar across cultures, indicating the presence of shared narrative templates.
Overall, our method reveals that current AI-generated stories primarily exhibits templated localization. 
This suggests that evaluations of cultural competence that do not account for the mechanism of localization may overestimate models' competence in generating  localized narratives and highlights the need for methods that capture deeper narrative variation.

\section{Limitations}
We evaluate models by providing a single prompt and evaluating the resultant generation.
The use of these systems in the real world might involve more complex interactions, like multi-turn conversations and detailed prompts \cite{walsh2025aifictionwild}.
An important direction of future research here is to understand the degree of detail in the prompt or during a multi-turn interaction that results in the model breaking out of its default homogeneous behavior, and the impact thereof on users from different sociocultural backgrounds.
Moreover, we focus on closed and open-sourced LLMs in a zero-shot prompting setting. We leave the examination of other types of specialized systems, such as \href{sudowrite.com}{Sudowrite}, a commercial software for fiction writers, tools with narrative planning \cite{xie_creating_2024}, or tools that are personalized for users or communities \cite{hamna_kahani_2024}, to future work.
Our method of measuring whether localization is templated will be useful to evaluate how more detailed prompts, stronger models, or other interventions impact the presence of templated localization in LLM generations.

We analyzed narratives generated when LLMs are instructed to write children's stories. 
While we based our selection of topics for these stories on established frame-works of cross-cultural variation in values, thus expecting that stories written for these topics may manifest these variations, we acknowledge that the genre of the narratives written can have an impact on the homogeneity. We hope that the community will utilize our framework to extend the evaluation to other genres of narratives, such as writing of screenplays, fiction for adults, essays, and even multimodal narratives like films.

We operationalize culture through the proxy of nationality. 
Future work must examine the homogenization effects at different levels of granularity within cultures, such as within a specific country \cite{bhagat_tales_2025} and for other axes of identities  \cite{seth_how_2025}. 

We relied on the SeeGULL dataset \cite{jha2023seegullstereotypebenchmarkbroad} as our source of reference stereotypical attributes. 
Since the candidate stereotypical attributes in SeeGULL were sourced from language models --- albeit different models than the ones evaluated here --- 
this may impact the degree of stereotypicality we observe. We chose this dataset for its broad coverage of nationalities and ratings obtained from raters residing in those countries. Future work should explore the use of different methods of collection of reference stereotypes, such as those created with community participation \cite{dev_building_2023}.

Finally, all stories we evaluated were generated in English, and we focused on examining the similarity in the narratives as operationalized through multi-word sequences. This was done to facilitate word-level analysis in identifying similarities to surface latent templates across stories.
An important direction of future research is to characterize the underlying narratives of these stories either computationally or manually through other forms of narrative representations such as discourse structure, themes, narrative events, characters, values, and so on \cite{hamilton_narrabench_2026,namuduri_qudsim_2025,Begus2024,piper_narrative_2021}

\section*{Acknowledgments}
We thank Anjali Kantharuban, Joel Mire, and Saujas Vaduguru for their feedback on early drafts of the manuscript. This work partially used computational resources from Bridges-2 \cite{bridges} at Pittsburgh Supercomputing Center through allocation CIS250960 from the Advanced Cyber infrastructure Coordination Ecosystem: Services \& Support (ACCESS) program, which is supported by National Science Foundation grants \#2138259, \#2138286, \#2138307, \#2137603, and \#2138296. This research was funded by the \href{https://ror.org/05xpvk416}{National Institute of Standards and Technology (NIST)} and the \href{https://ror.org/05x2bcf33}{Carnegie Mellon University AI Measurement Science and Engineering Center (AIMSEC)}. Shaily Bhatt (ORCID: 0000-0001-9616-6264) and Fernando Diaz (ORCID: 0000-0003-2345-1288) were funded by NIST through Federal Award ID Number 60NANB24D231.
\bibliography{custom}

\appendix
\section{NPMI Operationalization}
\label{app:cultural-markers}
We count the number of times that a word appears in stories of a specific culture and in stories across all cultures.  A higher positive NPMI indicates higher association between the word and the culture. NPMI has been used in measuring association between vocabulary and different classes within the corpora (such as specific communities) in prior work \cite{zhang_community_2017,lucy_words_2023,bhatt_research_2025}.

\section{Stereotypicality in Cultural Markers}
\label{app:stereotype-method}
We measure the stereotypicality of cultural markers by calculating the precision between the set of cultural markers ($\lexiconck{\culture}$) and the reference set of stereotypical attributes ($\stereotypesc{\culture}$). Specifically, precision is calculated as follows (where $|A|$ denotes the length of the set $A$):
\begin{align}
    \frac{|\lexiconck{\culture} ~ \cap ~ \stereotypesc{\culture}|}{|\lexiconck{\culture}|}
\end{align}

We further measure the average offensiveness by measuring the average offensiveness of the lexical items in the intersection set. Specifically, let $O(x)$ be the offensiveness score for a stereotypical attribute x, we calculate average offensiveness as:

\begin{align}
    \frac{\sum_{x \in \lexiconck{\culture} ~ \cap ~ \stereotypesc{\culture}} O(x)}{|\lexiconck{\culture} ~ \cap ~ \stereotypesc{\culture}|}
\end{align}

\section{Classifier Parameters}
\label{app:classifier-details}

We fine-tuned the mmBERT model \cite{marone2025mmbertmodernmultilingualencoder} to predict the culture given a story using our corpus. We fine-tuned the model with a classifier head for 193-way classification.
All classifiers were trained with total epochs set to 50 and an early stopping criterion of 5 epochs. The validation macro \Fone was used to stop training early if necessary. Each classifier was trained on 60\% of the stories, and 20\% of the stories were used for validation and testing, each. The max length of the model was set to 768 and a batch size of 32 was used. After preliminary trials, the learning rate was set to $3e^{-5}$ with a warm-up schedule of 500 steps. Random seed of 47 was used for reproducibility.  Table \ref{tab:classifier_details} shows the validation accuracy and best epoch for each of the final classifier.

\begin{table}[h]
\centering
\footnotesize
\begin{tabular}{lccc}
\toprule
LLM & Fold & Best Val Acc. & Epochs \\
\midrule
gemma-3-12b-instruct & 1 & 0.9050 & 14.0 \\
gemma-3-12b-instruct & 2 & 0.9074 & 22.0 \\
gemma-3-12b-instruct & 3 & 0.9069 & 26.0 \\
gemma-3-12b-instruct & 4 & 0.9053 & 15.0 \\
gemma-3-12b-instruct & 5 & 0.9017 & 10.0 \\
\midrule
gpt-3-5-turbo & 1 & 0.9659 & 10.0 \\
gpt-3-5-turbo & 2 & 0.9659 & 10.0 \\
gpt-3-5-turbo & 3 & 0.9660 & 9.0 \\
gpt-3-5-turbo & 4 & 0.9649 & 8.0 \\
gpt-3-5-turbo & 5 & 0.9665 & 10.0 \\
\midrule
gpt-4o-mini & 1 & 0.9691 & 8.0 \\
gpt-4o-mini & 2 & 0.9684 & 14.0 \\
gpt-4o-mini & 3 & 0.9692 & 14.0 \\
gpt-4o-mini & 4 & 0.9687 & 15.0 \\
gpt-4o-mini & 5 & 0.9688 & 16.0 \\
\midrule
llama-3-1-8B-instruct & 1 & 0.8080 & 8.0 \\
llama-3-1-8B-instruct & 2 & 0.8036 & 7.0 \\
llama-3-1-8B-instruct & 3 & 0.8094 & 7.0 \\
llama-3-1-8B-instruct & 4 & 0.8073 & 7.0 \\
llama-3-1-8B-instruct & 5 & 0.8089 & 7.0 \\
\midrule
llama-3-3-70b-instruct & 1 & 0.9390 & 9.0 \\
llama-3-3-70b-instruct & 2 & 0.9388 & 29.0 \\
llama-3-3-70b-instruct & 3 & 0.9398 & 15.0 \\
llama-3-3-70b-instruct & 4 & 0.9384 & 8.0 \\
llama-3-3-70b-instruct & 5 & 0.9382 & 8.0 \\
\bottomrule
\end{tabular}
\caption{Best validation accuracy and corresponding epoch for each LLM and fold.}
\label{tab:classifier_details}
\end{table}

\section{Complete Classifier Results}
\label{app:classifier-results}
\begin{figure*}
    \centering
    \begin{subfigure}{\columnwidth}
        \includegraphics[width=\linewidth]{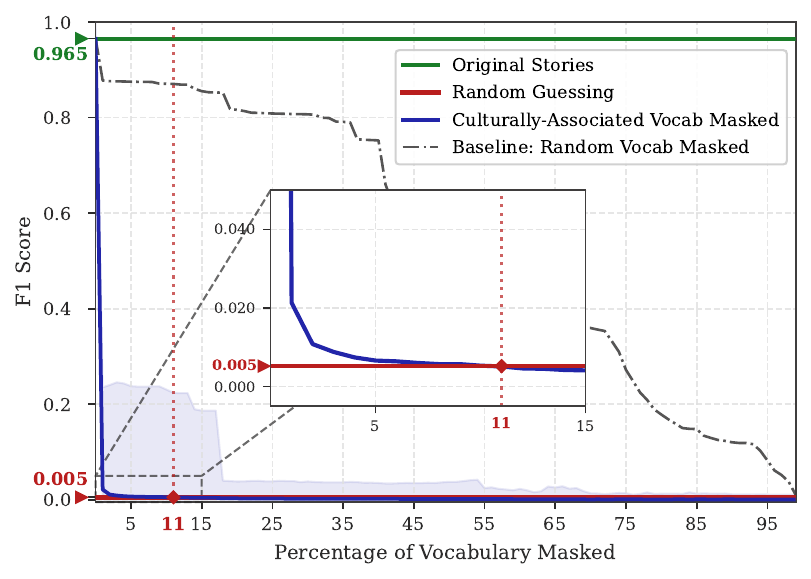}
        \caption{GPT 3.5 Turbo}
    \end{subfigure}
    \hfill
    \begin{subfigure}{\columnwidth}
        \includegraphics[width=\linewidth]{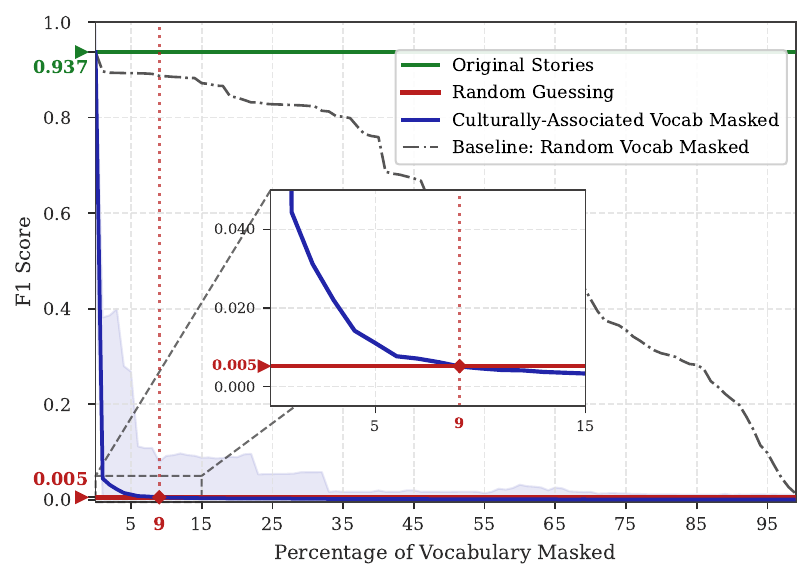}
        \caption{Llama 3.3 70B Instruct}
    \end{subfigure}

    \begin{subfigure}{\columnwidth}
        \includegraphics[width=\linewidth]{artifacts/graphs/macro_f1_llama_3-3_70b_instruct_classifier_with_nationality_template_with_nationality.pdf}
        \caption{Gemma 3 12B Instruct}
    \end{subfigure}
    \hfill
    \begin{subfigure}{\columnwidth}
        \includegraphics[width=\linewidth]{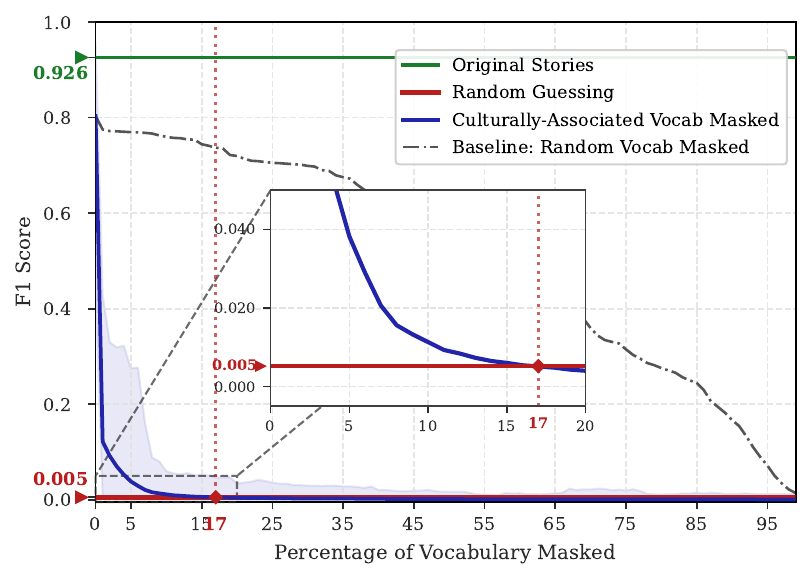}
        \caption{Llama 3.1 8B Instruct}
    \end{subfigure}

    \caption{Identifying Cultural markers. \Fone of nationality classifier as a function of number of masked words.}\label{app:fig:cultural-markers}
\end{figure*}

Figure \ref{app:fig:cultural-markers} shows the performance of the culture classifier for the remaining 4 LLMs. Similar to results described in \S~5.1, for all LLMs, when the culturally associated vocabulary is masked the classifier performance drops sharply. The drop is much more uniform for random masking. For all models, the performance on the original stories is near perfect.

\section{Complete Homogeneity Results}
\begin{table*}[t]
\centering
\small
\begin{tabular}{lccccccccc}
\toprule
 & \multicolumn{3}{c}{inter-group} & \multicolumn{3}{c}{intra-group} & \multicolumn{3}{c}{inter/intra} \\
\cmidrule(lr){2-4} \cmidrule(lr){5-7} \cmidrule(lr){8-10}
 & original & masked & \% inc. & original & masked & \% inc. & original & masked & \% inc. \\
\midrule
\multicolumn{10}{c}{Longest Common Subsequence} \\
\midrule
gpt 3.5 turbo & 0.0252 & 0.0170 & -32.39\% & 0.0347 & 0.0350 & 0.86\% & 0.7262 & 0.4857 & -33.1\% \\
gpt 4o mini & 0.0135 & 0.0097 & -27.90\% & 0.0172 & 0.0174 & 1.09\% & 0.7849 & 0.5575 & -29.0\% \\
gemma 3.12b instruct & 0.0114 & 0.0096 & -15.71\% & 0.0137 & 0.0148 & 7.99\% & 0.8321 & 0.6486 & -22.1\% \\
llama 3.1 8b instruct & 0.0122 & 0.0094 & -22.31\% & 0.0131 & 0.0134 & 2.84\% & 0.9313 & 0.7015 & -24.7\% \\
llama 3.3 70b instruct & 0.0180 & 0.0134 & -25.38\% & 0.0263 & 0.0265 & 0.56\% & 0.6844 & 0.5057 & -26.1\% \\
\midrule
\multicolumn{10}{c}{Jaccard (4-gram shingles)} \\
\midrule
gpt 3.5 turbo & 0.0299 & 0.0141 & -53.02\% & 0.0419 & 0.0426 & 1.71\% & 0.7136 & 0.3310 & -53.6\% \\
gpt 4o mini & 0.0161 & 0.0080 & -49.93\% & 0.0217 & 0.0224 & 3.36\% & 0.7419 & 0.3571 & -51.9\% \\
gemma 3.12b instruct & 0.0075 & 0.0044 & -40.64\% & 0.0122 & 0.0176 & 43.59\% & 0.6148 & 0.2500 & -59.3\% \\
llama 3.1 8b instruct & 0.0093 & 0.0040 & -56.74\% & 0.0110 & 0.0123 & 11.77\% & 0.8455 & 0.3252 & -61.5\% \\
llama 3.3 70b instruct & 0.0218 & 0.0120 & -44.76\% & 0.0339 & 0.0345 & 1.55\% & 0.6431 & 0.3478 & -45.9\% \\
\bottomrule
\end{tabular}
\caption{Narrative homogeneity.  Multi-word similarity amongst stories on the same topic in either their original form or when random words equivalent to the number of cultural markers are masked.   Intra-group measures the similarity amongst stories for the same nationality.  The last column group divides the inter-group similarity by the intra-group similarity to control for similarity attributable to a cross-nationality template.  }
\label{tab:app-homogeneity}
\end{table*}
Table \ref{tab:app-homogeneity} shows the homogeneity, as measured by average similarity of stories when random words equivalent to the number of cultural markers are masked. 

\section{Complete Stereotyping Results}
\label{app:stereotype-results}
\begin{table}[]
\footnotesize
\begin{tabular}{lrlr}
\toprule
Nationality  & Precision & Nationality  & Offensiveness \\ \midrule
Australia   & 0.0172                        & Rwanda      & 4                               \\
India       & 0.0095                        & Bosnia      & 3                               \\
Mexico      & 0.0093                        & Bulgaria    & 3                               \\
New Zealand & 0.0088                        & Honduras    & 2.667                           \\
Japan       & 0.0082                        & Gabon       & 2.667                           \\
France      & 0.0077                        & Indonesia   & 2.667                           \\
Italy       & 0.0077                        & Sri Lanka   & 2                               \\
Ethiopia    & 0.0068                        & Pakistan    & 1.75                            \\
America     & 0.0064                        & Namibia     & 1.667                           \\
Vietnam     & 0.0054                        & Colombia    & 1.667                           \\  \bottomrule
\end{tabular}
\caption{Stereotypes. Top 10 countries with highest stereotype precision and offensiveness as rated by regional raters for cultural markers form GPT 3.5 Turbo.}
\end{table}

\begin{table}[]
\footnotesize
\begin{tabular}{lrlr}
\toprule
Nationality  & Precision & Nationality  & Offensiveness \\ \midrule
Australia   & 0.0124                        & Tunisia       & 4                               \\
America     & 0.0083                        & Rwanda        & 3.333                           \\
China       & 0.0077                        & Bosnia        & 3                               \\
India       & 0.0056                        & Colombia      & 2.667                           \\
Germany     & 0.0055                        & Liberia       & 2.5                             \\
New Zealand & 0.0055                        & Bangladesh    & 2.333                           \\
Italy       & 0.0049                        & Gambia        & 2                               \\
Japan       & 0.0046                        & Rep. of Congo & 2                               \\
North Korea & 0.0044                        & Afghanistan   & 1.389                           \\
Pakistan    & 0.0036                        & Mexico        & 1.333                           \\    \bottomrule
\end{tabular}
\caption{Stereotypes. Top 10 countries with highest stereotype precision and offensiveness as rated by regional raters for cultural markers form Llama 3.1 8B Instruct.}
\end{table}

\begin{table}[]
\footnotesize
\begin{tabular}{lrlr}
\toprule
Nationality  & Precision & Nationality  & Offensiveness \\ \midrule
Australia   & 0.0175                        & Rwanda      & 4                               \\
China       & 0.0074                        & Gambia      & 3                               \\
Italy       & 0.0073                        & Cameroon    & 2.667                           \\
Japan       & 0.0061                        & Venezuela   & 2.667                           \\
Vietnam     & 0.0059                        & Bangladesh  & 2.667                           \\
North Korea & 0.0059                        & Poland      & 2.667                           \\
New Zealand & 0.0056                        & Burundi     & 2.333                           \\
Greece      & 0.0055                        & Mexico      & 2                               \\
America     & 0.0055                        & Argentina   & 2                               \\
Ethiopia    & 0.0045                        & Pakistan    & 1.556                           \\   \bottomrule
\end{tabular}
\caption{Stereotypes. Top 10 countries with highest stereotype precision and offensiveness as rated by regional raters for cultural markers form Llama 3.3 70B Instruct.}
\end{table}

\begin{table}[]
\footnotesize
\begin{tabular}{lrlr}
\toprule
Nationality  & Precision & Nationality  & Offensiveness \\ \midrule
Australia   & 0.0121                        & Yemen       & 4                               \\
India       & 0.0091                        & Bosnia      & 4                               \\
Japan       & 0.0079                        & Angola      & 3.667                           \\
China       & 0.0078                        & Iraq        & 3.667                           \\
Afghanistan & 0.0075                        & Algeria     & 3.333                           \\
North Korea & 0.0058                        & Romania     & 3.333                           \\
New Zealand & 0.0056                        & Azerbaijan  & 3.333                           \\
America     & 0.0046                        & Syria       & 3                               \\
Pakistan    & 0.0044                        & Cameroon    & 3                               \\
Spain        & 0.0037                        & Albania     & 2.667                           \\  \bottomrule
\end{tabular}
\caption{Stereotypes. Top 10 countries with highest stereotype precision and offensiveness as rated by regional raters for cultural markers form Gemma 3 12B Instruct.}
\end{table}

Tables 6 to 9 show the top 10 countries with highest stereotype precision and offensiveness for all models. We see patterns similar to those observed in regional results for GPT 4o Mini stories in \S~5.3. Table \ref{tab:stereotype-eg} shows examples of stereotypical attributes found in the cultural markers, as calculated using the SeeGULL dataset.

\begin{table*}[]
    \centering
    \footnotesize
    \begin{tabular}{ll}
    \toprule
    Nationality  & Stereotypes                                                 \\
    \midrule
    \multicolumn{2}{c}{GPT 4o Mini}                                            \\
    \midrule
    Syria        & violent                                                     \\
    China        & witty, sophisticated                                        \\
    Bangladesh   & beggar                                                      \\
    Gambia       & unreliable                                                  \\
    India        & mystic, religious, witty, vegetarian, untidy, undisciplined \\
    \midrule
    \multicolumn{2}{c}{GPT 3.5 Turbo}                                          \\
    \midrule
    Denmark      & progressive                                                 \\
    Afghanistan  & backward, killer                                            \\
    Iran         & oppressive                                                  \\
    Pakistan     & criminal, paranoid, uneducated                              \\
    Turkey       & ruthless, conservative                                      \\
    \midrule
    \multicolumn{2}{c}{Llama 3.1 8B Instruct}                                  \\
    \midrule
    Cameroon     & tribal, talkative                                           \\
    German       & punctual, thorough, blonde                                  \\
    Italy        & expressive, temperamental, foodie                           \\
    Kenya        & industrious                                                 \\
    Liberia      & barbaric, uneducated                                        \\
    \midrule
    \multicolumn{2}{c}{Llama 3.3 70B Instruct}                                 \\
    \midrule
    Vietnam      & smelly, communist                                           \\
    Japan        & ninja, courteous, samurai                                   \\
    Nepal        & rational                                                    \\
    Nigeria      & smelly, witch                                               \\
    South Africa & apartheid, unfriendly                                       \\
    \midrule
    \multicolumn{2}{c}{Gemma 3 12B Instruct}                                   \\
    \midrule
    Lebanon      & terrorist                                                   \\
    Uganda       & untrustworthy                                               \\
    Britain      & proper, aloof                                               \\
    Venezuela    & dumb                                                        \\
    Israel       & pushy                                                      \\
    \bottomrule
    \end{tabular}
    \caption{Example of cultural markers that overlap with stereotypes from SeeGULL dataset.}
    \label{tab:stereotype-eg}
    \end{table*}

\end{document}